\documentclass[sigconf]{acmart}
\acmConference[ ]{ }{ }{ }
\AtBeginDocument{%
  }
\setcopyright{acmlicensed}
\copyrightyear{2026}
\acmYear{2026}
\acmConference[HRI '26]{IEEE/ACM International Conference on Human-Robot Interaction}{March 16--19, 2026}{Edinburgh, Scotland, UK}
\acmBooktitle{IEEE/ACM International Conference on Human-Robot Interaction (HRI '26), March 16--19, 2026, Edinburgh, Scotland, UK}
\usepackage{romannum}
\usepackage{float}
\usepackage{subcaption}  
\begin{document}
\title{HumanDiffusion: A Vision-Based Diffusion Trajectory Planner with Human-Conditioned Goals for Search and Rescue UAV}
\author{Faryal Batool}
\affiliation{%
  \institution{Skolkovo Institute of Science and Technology}
  \city{Moscow}
  \country{Russia}}
\email{Faryal.Batool@skoltech.ru}

\author{Iana Zhura}
\authornote{Equal contribution.}
\affiliation{%
  \institution{Skolkovo Institute of Science and Technology}
  \city{Moscow}
  \country{Russia}}
\email{iana.zhura@skoltech.ru}

\author{Valerii Serpiva}
\authornotemark[1]
\affiliation{%
  \institution{Skolkovo Institute of Science and Technology}
  \city{Moscow}
  \country{Russia}}
\email{valerii.serpiva@skoltech.ru}

\author{Roohan Ahmed Khan}
\authornotemark[1]
\affiliation{%
  \institution{Skolkovo Institute of Science and Technology}
  \city{Moscow}
  \country{Russia}}
\email{ra.khan@skoltech.ru}

\author{Ivan Valuev}
\affiliation{%
  \institution{Skolkovo Institute of Science and Technology}
  \city{Moscow}
  \country{Russia}}
\email{ivan.valuev@skoltech.ru}

\author{Issatay Tokmurziyev}
\affiliation{%
  \institution{Skolkovo Institute of Science and Technology}
  \city{Moscow}
  \country{Russia}}
\email{Issatay.Tokmurziyev@skoltech.ru}

\author{Dzmitry Tsetserukou}
\affiliation{%
  \institution{Skolkovo Institute of Science and Technology}
  \city{Moscow}
  \country{Russia}}
\email{d.tsetserukou@skoltech.ru}
\renewcommand{\shortauthors}{Batool et al.}
\begin{teaserfigure}
    \centering
    \includegraphics[width=0.9\textwidth,height=6cm,
        keepaspectratio]{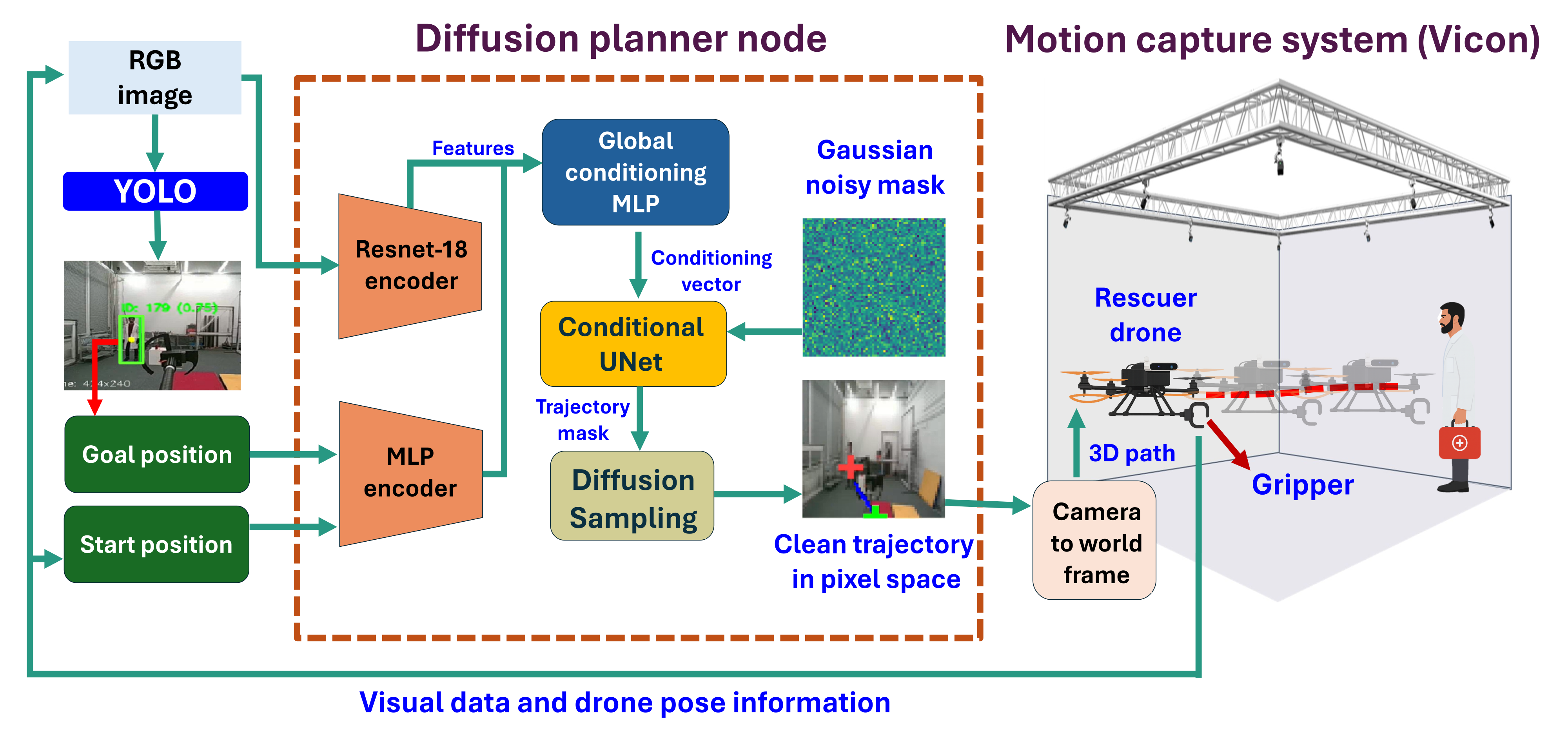}
    \caption{
    HumanDiffusion architecture. YOLO provides human-based goal points, then the image and start–goal information are encoded and fused to condition a UNet-based diffusion model. The model generates a clean pixel-space trajectory, which is converted to a 3D world-frame path for execution by the rescuer drone with a gripper.
    }
    \label{fig:system-architecture}
\end{teaserfigure}
\begin{abstract}
Reliable human--robot collaboration in emergency scenarios requires autonomous systems that can detect humans, infer navigation goals, and operate safely in dynamic environments. This paper presents \textbf{HumanDiffusion}, a lightweight image-conditioned diffusion planner that generates human-aware navigation trajectories directly from RGB imagery. The system combines YOLO-11--based human detection with diffusion-driven trajectory generation, enabling a quadrotor to approach a target person and deliver medical assistance without relying on prior maps or computationally intensive planning pipelines. Trajectories are predicted in pixel space, ensuring smooth motion and a consistent safety margin around humans.
We evaluate HumanDiffusion in simulation and real-world indoor mock-disaster scenarios. On a 300-sample test set, the model achieves a mean squared error of 0.02 in pixel-space trajectory reconstruction. Real-world experiments demonstrate an overall mission success rate of \textbf{80\%} across accident-response and search-and-locate tasks with partial occlusions. These results indicate that human-conditioned diffusion planning offers a practical and robust solution for human-aware UAV navigation in time-critical assistance settings.
\end{abstract}


\maketitle
\section{Introduction}
Search and rescue (SAR) missions often operate under severe time constraints and limited situational awareness, where the locations of victims or medical personnel are unknown. Unmanned aerial vehicles (UAVs) are well suited to such settings due to their ability to access confined or hazardous environments and provide rapid assistance \cite{goodrich2008supporting}. While modern SAR systems commonly integrate vision-based human detection \cite{wang2023yolo,giusti2015machine} with autonomous navigation, most rely on predefined goals, explicit maps, or planning frameworks with significant computational overhead such as A*, RRT*, or Model Predictive Control (MPC) \cite{köhler2025mpcframeworkefficientnavigation,MPC_RRT}. These assumptions limit applicability in dynamic or partially observable environments where human locations must be inferred online.
We propose a lightweight \textit{human-conditioned diffusion planner} that generates global trajectories directly from RGB images, leveraging YOLO-based human detector. The center of the detected human bounding box is treated as an implicit goal, eliminating the need for maps and waypoints in goal inference. Although designed for real-world SAR, we evaluate the system in controlled indoor environments as a proof-of-concept, focusing on medical handover tasks such as retrieving supplies from one individual and delivering them to another.

Our results show that diffusion-based planners can work well as perception-driven global planners. This provides a basis for future work that combines local obstacle avoidance and supports scalable human–robot collaboration in emergency response. The main contributions of this paper are:
\begin{itemize}
    \item \textbf{End-to-end image-conditioned diffusion planning:} We develop a lightweight diffusion model that generates global trajectories conditioned solely on RGB images and the inferred start--goal pair, enabling map-free navigation.
    
    \item \textbf{Integrated proof-of-concept system:} We implement a fully functional indoor UAV pipeline supporting person identification, human-based goal generation, and autonomous trajectory execution with a gripper for object handover.
    
    \item \textbf{Sim-to-real deployment:} We train a diffusion planner solely on simulated RGB data and A*-generated trajectories and successfully deploy it in two real-world indoor assistance scenarios.
    

\end{itemize}

\section{Related Works}
This section reviews prior work in UAV-based SAR, human-centered navigation, classical and learning-based UAV planning, diffusion-based trajectory generation, and multimodal vision--language UAV systems.

\subsection{Human-Centered UAV Navigation for Search and Rescue}
Prior UAV-SAR research emphasizes robust human perception, with YOLO-based detectors widely adopted for detecting small, distant, or partially occluded humans \cite{ciccone2025yolosar,ramirez2023wooded,abbas2024humanrecog}. Surveys further highlight human detection as a core capability for aerial SAR platforms \cite{lyu2023uavsurvey,zhang2025aerialperson}. Several works combine detection with filtering-based tracking to enable UAV-based human following \cite{chhabra2024humanfollowinguav,mueller2024indoorhumanfollow}. However, these approaches assume persistent visual contact and focus on local interaction, without connecting human perception to global navigation or planning. In contrast, our approach directly uses human detection to infer navigation goals and generate global trajectories.

\subsection{UAV Navigation and Planning}
Conventional UAV navigation pipelines rely on occupancy maps, classical planners such as A* or RRT*, and control strategies including MPC or learning-based controllers such as Reinforcement Learning for trajectory execution \cite{gaigalas2025astarvision,baidya2024landing,kucukerdem2025autocontrol,MPC_RRT,ego_planner,agile_pilot,marlander}. While effective in structured environments, these approaches typically depend on explicit maps, predefined waypoints, or accurate state estimation, which limits their applicability in scenarios where only onboard vision is available and navigation goals must be inferred online from perception.

\subsection{Diffusion Models for Trajectory Planning}
Diffusion models have recently gained significant attention for motion planning and trajectory synthesis. Vision-based diffusion planners such as NoMaD \cite{vuckovic2024diffnav} and DiPPeR \cite{DiPPeR} generate navigation trajectories conditioned on RGB observations, cost fields, or object-level goals, demonstrating strong generalization in unknown and cluttered environments compared to classical planners. Recent surveys further highlight the rapid adoption of diffusion models for robotic motion generation and decision-making \cite{wolf2025manipsurvey,ubukata2024planningreview,swarmdiffusionen}.

\subsection{Vision--Language Models for UAV Reasoning}
Recent work integrates vision--language models (VLMs) with UAVs to enable semantic reasoning and high-level mission understanding. UAV-VLRR combines VLM-based scene interpretation with NMPC for SAR tasks \cite{uavvlrr2025}, while FlightGPT and UAV-VLA explores language-guided UAV navigation \cite{flightgpt2025,UAV_VLA}. Other studies investigate VLM-guided object detection and navigation \cite{navblip2025,openvocabvlm2025,aienhanced2025}. Despite their strong reasoning capabilities, these systems do not couple human-aware perception with generative trajectory planning or automatic goal inference.
Overall, prior methods either detect humans without planning, perform local following, rely on predefined goals, or employ diffusion planners without human-conditioned goal inference. Our work bridges this gap by unifying human detection with diffusion-based trajectory generation, enabling end-to-end, map-free global planning for search and rescue, driven directly by detected humans.

\section{System Architecture}

The HumanDiffusion framework as shown in Figure ~\ref{fig:system-architecture} consists of two core modules: (i) a YOLO-based perception system that detects humans and outputs a dynamic goal point, and (ii) a diffusion-based trajectory generator that predicts a path in pixel space conditioned on the RGB image, start and inferred goal location.

\subsection{Perception and Human Goal Inference}

The YOLO-11 detector identifies humans from incoming RGB frames, and the center of the selected bounding box is used as the navigation goal. This goal updates continuously, allowing the UAV to track a moving human, while the start point is obtained from onboard localization and provided to the diffusion model as an additional conditioning signal.

\subsection{Diffusion-Based Trajectory Generator}

The trajectory generator is based on a conditional UNet-based diffusion model inspired by \cite{liang2024dtg}. The model predicts a pixel-space trajectory mask by iteratively denoising a noisy sample. The input is a three-channel mask  
$x_0 \in \mathbb{R}^{B \times 3 \times H \times W}$,  
where $B$ is batch size, $H$ and $W$ are spatial dimensions, and the channels represent the start point, goal point, and trajectory mask.

\paragraph{Forward Diffusion Process.}
Noise is gradually added to the clean mask using a squared-cosine schedule:
\[
x_t = \sqrt{\overline{\alpha}_t}\, x_0 + \sqrt{1 - \overline{\alpha}_t}\, \epsilon,
\]
where $x_t$ is the noisy sample at timestep $t$,  
$\overline{\alpha}_t = \prod_{s=1}^{t} \alpha_s$ is the cumulative noise factor,  
$\alpha_t = 1 - \beta_t$ with $\beta_t$ the noise variance,  
and $\epsilon \sim \mathcal{N}(0,I)$ is the Gaussian noise.

\paragraph{Reverse Denoising Process.}
During denoising, the conditional UNet predicts the clean mask $\hat{x}_0$, which is used in the Denoising Diffusion Probabilistic Model (DDPM) posterior:
\[
x_{t-1} = \mu_t(x_t,\hat{x}_0) + \sigma_t z, \quad z \sim \mathcal{N}(0,I),
\]
where $\mu_t$ and $\sigma_t$ are the posterior mean and standard deviation and and $z \sim \mathcal{N}(0,I)$ is the Gaussian noise. This process is repeated from $t=T$ to $t=0$ while the start and goal channels are inpainted to ensure that the generated trajectory remains aligned with the specified boundary conditions.

\paragraph{Training Objective.}
The model is trained to reconstruct the trajectory mask and enforce accurate endpoints. The total loss is:
\[
\mathcal{L} = \lambda_{\text{path}}\,L_{\text{path}}
            + \lambda_{\text{endpoint}}\,L_{\text{endpoint}},
\]
where $\lambda_{\text{path}}$ and $\lambda_{\text{endpoint}}$ are the weights for trajectory reconstruction and endpoint accuracy respectively.

The trajectory reconstruction loss is:
\[
L_{\text{path}}
=
w_t \frac{1}{N}
\sum_{B,H,W}
\left(T^{\text{pred}}_{B,H,W} - T^{\text{gt}}_{B,H,W}\right)^2,
\]
where $T^{\text{pred}}$ and $T^{\text{gt}}$ are the predicted and ground-truth masks, respectively,  
$w_t$ is the trajectory-channel weight,  
and and $N = B \times H \times W$ is the total
number of pixels across the batch. The endpoint loss is:
\[
L_{\text{endpoint}} = \frac{1}{2}\left[
\begin{aligned}
 &w_s \frac{1}{N} \sum_{B,H,W}
  \left(S^{\text{pred}}_{B,H,W} - S^{\text{gt}}_{B,H,W}\right)^2
 \\[4pt]
 &+\quad\quad
 w_g \frac{1}{N} \sum_{B,H,W}
   \left(G^{\text{pred}}_{B,H,W} - G^{\text{gt}}_{B,H,W}\right)^2
\end{aligned}
\right].
\]
where $S^{\text{pred}}, G^{\text{pred}}$ and $S^{\text{gt}}, G^{\text{gt}}$ denote predicted and ground-truth start and goal masks, respectively, with channel weights $w_s$ and $w_g$. The factor $\tfrac{1}{2}$ normalizes their combined contribution.
The path loss ensures global geometric fidelity of the predicted trajectory, whereas the endpoint loss enforces accurate boundary conditions. 
\subsection{Training Pipeline and Dataset Generation}
We generate 9{,}800 ground-truth trajectories using the A* planner on simulated environments from~\cite{KennyLHW_VLNGo2Matterport_2024}. Among these, 8{,}000 samples are used for training, 1{,}500 for evaluation during training without doing any backpropagation, and 300 for testing. The dataset spans multiple indoor scenarios and start–goal pairs to promote generalization. Across multiple configurations, training with 100 diffusion steps and 30 epochs yielded the best performance.
\section{Experimental Evaluation}
The proposed HumanDiffusion pipeline was evaluated in real-world trials using a custom-built quadrotor equipped with an Intel RealSense~D455 depth camera and an Intel NUC for onboard computation. The YOLO-11 detector identifed humans and generated dynamically updated goal points, while the diffusion-based planner produced pixel-space trajectories at rate of 0.2--0.3\,s per frame. Both models ran off-board on a remote server and communicated with the onboard NUC via ROS. A custom gripper was attached for payload handling.
The diffusion model outputs 2D trajectories in image pixel space, which are projected into 3D waypoints using depth measurements and calibrated camera intrinsics. No explicit obstacle map is constructed; instead, the planner operates vision-driven, mapless navigation framework.

\subsection{Results and Failure Analysis}
The system was evaluated on a simulated test dataset and two real-world human-assistance scenarios: (1) \textit{Accident Response} and (2) \textit{Search-and-Locate in Occluded Environments}. Each scenario was executed 10 times, resulting in an overall success rate of 80\%.
Failures were categorized as follows: (i) \textit{Perception loss} (2 trials), caused by camera limitations or severe human occlusion; (ii) \textit{Controller tracking errors} (1 trial), due to transient state-estimation drift; and (iii) \textit{Communication dropouts} (1 trial), which delayed trajectory updates.

All experiments were conducted in accordance with laboratory safety and ethical guidelines. The UAV was equipped with propeller guards, operated at a low flight speed of 0.3\,m/s, and maintained a fixed stopping distance of 1\,m from participants. No physical contact occurred during the trials. All three participants provided informed consent prior to experimentation.

\subsection{Evaluation on Test Dataset}
\begin{figure}[t]
\centering
\includegraphics[width=1.0\linewidth]{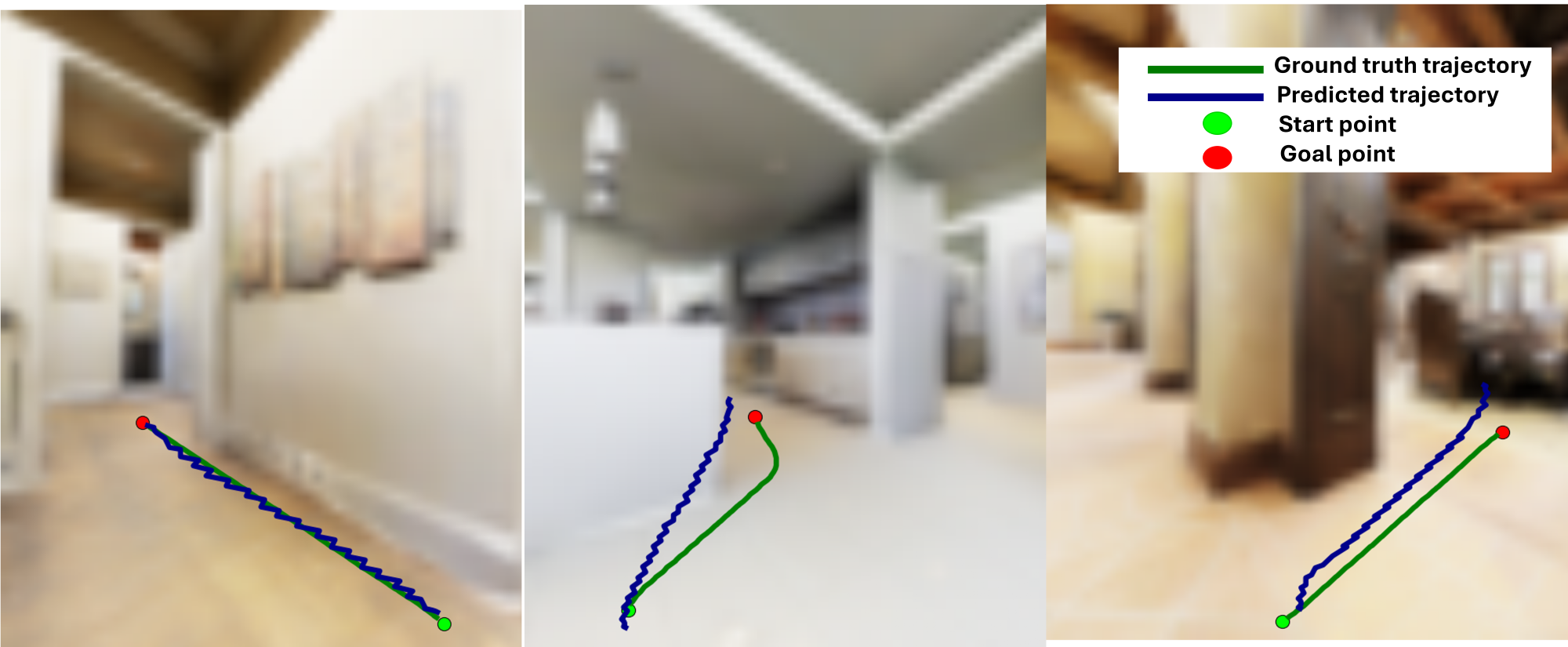}
\caption{
Comparison between diffusion predicted and annotated ground truth trajectories.
}
\label{fig:test_dataset}
\end{figure}

The model was first evaluated on a test dataset comprising of 300 simulated images with corresponding ground-truth trajectories. Performance was measured using the mean squared error (MSE) between predicted and ground-truth trajectory masks in pixel space. The evaluation yielded an MSE of \textbf{0.02}, indicating close agreement between predicted and reference trajectories. Representative qualitative results are shown in Figure~\ref{fig:test_dataset}. Minor zigzag artifacts appear in the predicted trajectories due to the original annotations being generated at a resolution of $512 \times 512$ and subsequently downsampled to $64 \times 64$ for training.

\subsection{Evaluation on Real Indoor Scenarios}
\subsubsection{Scenario 1: Accident Response}
\begin{figure}[t]
\centering
\includegraphics[width=1.0\linewidth]{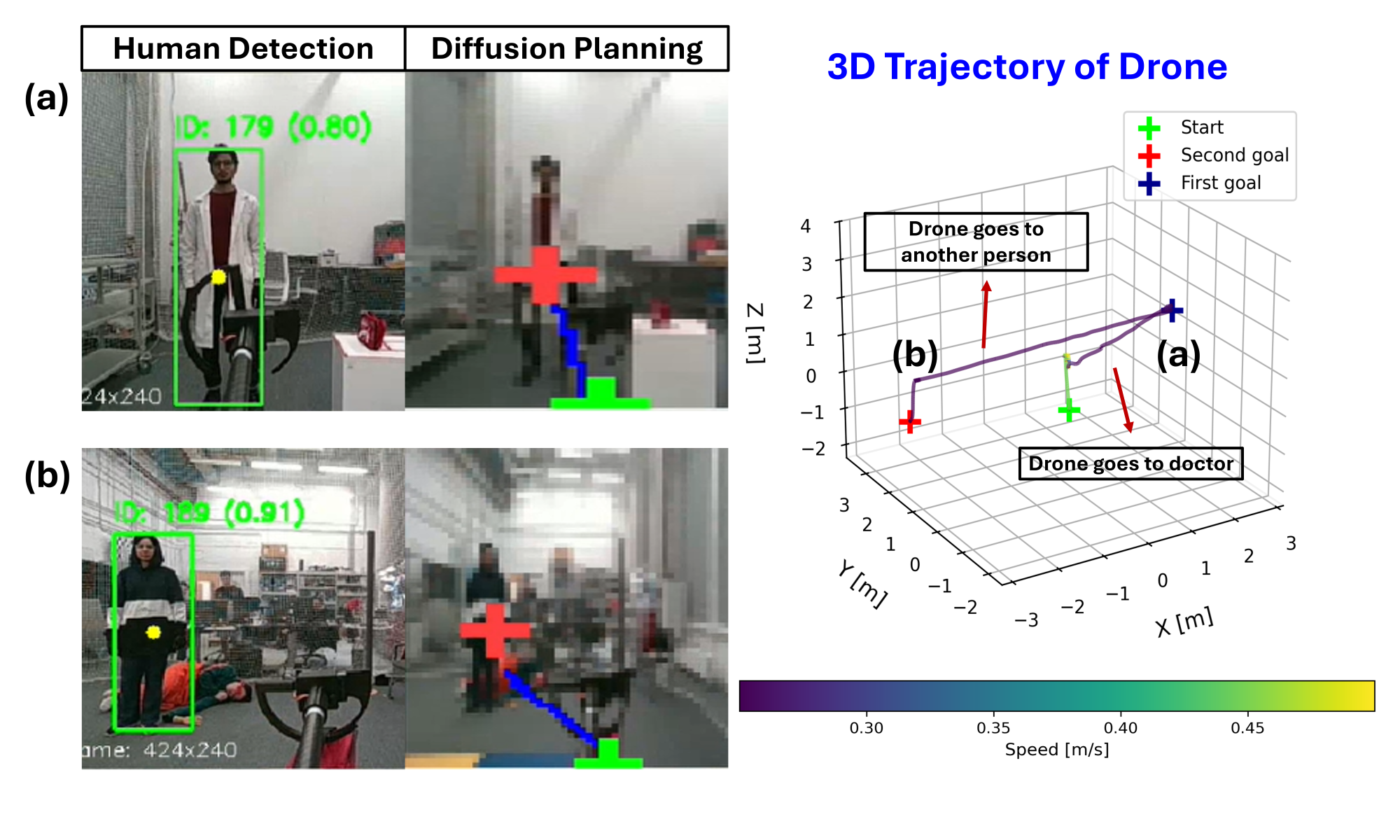}
\caption{
Human detection results for Scenario~1 with the corresponding diffusion-planned trajectories and the executed 3D flight path showing start position and goal updates.
}
\label{fig:accident_scenario}
\end{figure}

In this scenario, the UAV is provided only with approximate locations of the hospital and accident site. Upon arriving at the hospital, the drone detects a doctor using YOLO-11, extracts the bounding-box center as the navigation goal, and approaches while respecting the safety margin. After receiving a medical kit, the UAV autonomously navigates to the accident site, identifies an injured person, and completes the handover.
Across 10 trials, the system successfully completed 9 full delivery cycles. Figure~\ref{fig:accident_scenario} illustrates a representative trajectory, including human detection and safe-distance stopping behavior.

\subsubsection{Scenario 2: Search-and-Locate in Occluded Environments}
This scenario evaluated the system’s ability to track a partially occluded human concealed behind obstacles such as small hills or vegetation. When the human temporarily left the camera’s field of view, the UAV continued navigation toward the last known goal position. Once visibility was restored, the goal was updated and the UAV completed delivery of water or medical supplies.

\begin{figure}[t]
\centering
\includegraphics[width=1.0\linewidth]{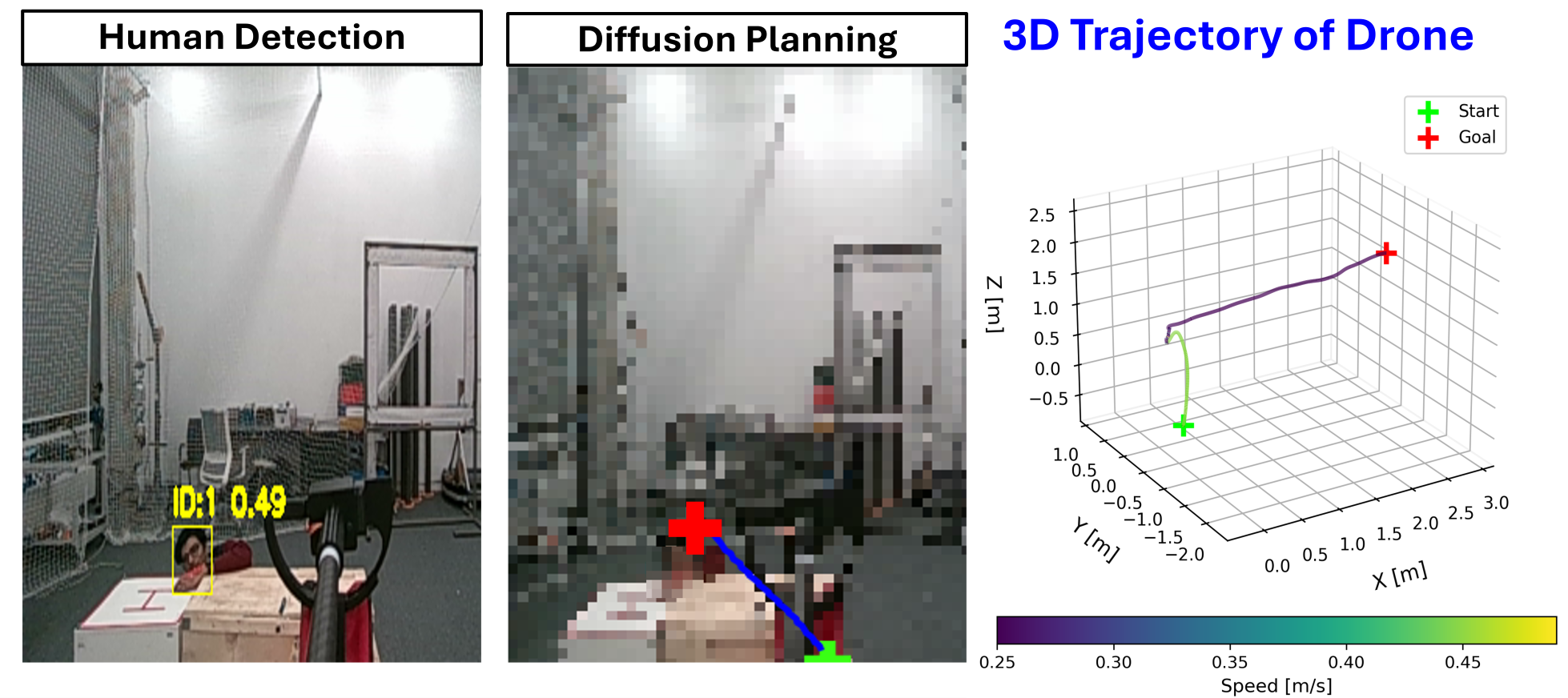}
\caption{
Human detection results for Scenario~2 with the corresponding diffusion-planned trajectories and the executed 3D flight path showing start position and goal updates.
}
\label{fig:forest_scenario}
\end{figure}

The system succeeded in 7 out of 10 trials. Figure~\ref{fig:forest_scenario} shows an example of successful target reacquisition following temporary occlusion.
Overall, HumanDiffusion achieved an 80\% success rate in real-world trials. Performance in Scenario~2 was slightly lower due to communication delays and limited arena size, where the UAV occasionally detected the human only after reaching the operational boundary, leaving insufficient space for redirection under the enforced 1\,m stopping margin. Despite these constraints, the results demonstrate that HumanDiffusion enables reliable, safe, and human-aware assistance through diffusion-based trajectory planning combined with real-time human detection.
\section{Conclusion and Future Work}
This work introduced HumanDiffusion, a vision based, diffusion driven trajectory generation framework for human assistance with autonomous aerial robots. By combining YOLO-11 based human detection with a conditional diffusion model, the system produces smooth, safe, and goal consistent trajectories directly in pixel space. Real-world evaluations showed reliable performance across accident response and search-and-locate scenarios, achieving an overall success rate of 80\%. These results indicate that diffusion based planning is a robust alternative to classical navigation methods in dynamic human–robot interaction settings.
Future work will extend HumanDiffusion with gesture guided human selection, multi human handling with prioritization and target switching as in \cite{zhou2019omniscalefeaturelearningperson,Rollo_2023}, and improved collision awareness in dynamic environments. These directions move the system toward fully autonomous, context aware aerial assistance suitable for real-world deployment.
\section*{Acknowledgment}
Research reported in this publication was financially supported by the RSF--DST grant No.~24-41-02039.

\bibliographystyle{ACM-Reference-Format}
\bibliography{bib}

@ARTICLE{chhabra2024humanfollowinguav,
  author={Bany Abdelnabi, Ahmad A. and Rabadi, Ghaith},
  journal={IEEE Access}, 
  title={Human Detection From Unmanned Aerial Vehicles’ Images for Search and Rescue Missions: A State-of-the-Art Review}, 
  year={2024},
  volume={12},
  number={},
  month = {Oct 14,},
  pages={152009--152035}}

@ARTICLE{goodrich2008supporting,
  author={Khosravi, Mohammadjavad and Arora, Rushiv and Enayati, Saeede and Pishro-Nik, Hossein},
  journal={IEEE Transactions on Automation Science and Engineering}, 
  title={A Search and Detection Autonomous Drone System: From Design to Implementation}, 
  year={2025},
  volume={22},
  number={},
 month = {May 08-10,},
  pages={3485--3501},}

@INPROCEEDINGS{wang2023yolo,
  author={V, Pavan Kumar and Reddy, G. Likhith and Mahadev, H. and Harish Kumar, V. and Prasad, K. Krishna},
  booktitle={Proc. IEEE Int. Conf. on Inventive Research in Computing Applications (ICIRCA)}, 
  title={Real-Time Detection of Unmanned Aerial Vehicles Using YOLOv8}, 
  year={2025},
  volume={},
  number={},
month = {June 25-27,},
  pages={216--223}}

@ARTICLE{giusti2015machine,
  author={Giusti, Alessandro and Guzzi, Jérôme and Cireşan, Dan C. and He, Fang-Lin and Rodríguez, Juan P. and Fontana, Flavio and Faessler, Matthias and Forster, Christian and Schmidhuber, Jürgen and Caro, Gianni Di and Scaramuzza, Davide and Gambardella, Luca M.},
  journal={IEEE Robotics and Automation Letters}, 
  title={A Machine Learning Approach to Visual Perception of Forest Trails for Mobile Robots}, 
  year={2016},
  volume={1},
  number={2},
month = {Dec. 17,},
  pages={661--667}}

@misc{köhler2025mpcframeworkefficientnavigation,
      title={An MPC framework for efficient navigation of mobile robots in cluttered environments}, 
      author={Johannes Köhler and Daniel Zhang and Raffaele Soloperto and Andrea Carron and Melanie Zeilinger},
      year={2025},
      note={arxiv:2509.15917}, 
}

@INPROCEEDINGS{MPC_RRT,
  author={Primatesta, Stefano and Pagliano, Alessandro and Guglieri, Giorgio and Rizzo, Alessandro},
  booktitle={Proc. IEEE Int. Conf. on Unmanned Aircraft Systems (ICUAS)}, 
  title={Model Predictive Sample-based Motion Planning for Unmanned Aircraft Systems}, 
  year={2021},
  volume={},
  number={},
month = {June 15-18,},
  pages={111--119}}

@article{lyu2023uavsurvey,
title = {Unmanned aerial systems in search and rescue: A global perspective on current challenges and future applications},
author = {Carlos Osorio Quero and Jose Martinez-Carranza},
journal = {International Journal of Disaster Risk Reduction},
volume = {118},
pages = {105199},
month = {Feb. 6,},
year = {2025},
}

@article{
zhang2025aerialperson,
author = {Xiangqing Zhang  and Yan Feng  and Nan Wang  and Guohua Lu  and Shaohui Mei },
title = {Aerial Person Detection for Search and Rescue: Survey and Benchmarks},
journal = {Journal of Remote Sensing},
volume = {5},
pages = {0474},
month = {March 25,},
year = {2025}}

@Article{ciccone2025yolosar,
AUTHOR = {Ciccone, Francesco and Ceruti, Alessandro},
TITLE = {Real-Time Search and Rescue with Drones: A Deep Learning Approach for Small-Object Detection Based on YOLO},
JOURNAL = {Drones},
VOLUME = {9},
MONTH = {July 22,},
YEAR = {2025},
NUMBER = {8},

}

@Article{ramirez2023wooded,
AUTHOR = {Ramírez-Ayala, Oscar and González-Hernández, Iván and Salazar, Sergio and Flores, Jonathan and Lozano, Rogelio},
TITLE = {Real-Time Person Detection in Wooded Areas Using Thermal Images from an Aerial Perspective},
JOURNAL = {Sensors},
VOLUME = {23},
MONTH = {Nov. 16,},
YEAR = {2023},
NUMBER = {22},
}

@ARTICLE{abbas2024humanrecog,
AUTHOR={Abbas, Yawar  and Al Mudawi, Naif  and Alabdullah, Bayan  and Sadiq, Touseef  and Algarni, Asaad  and Rahman, Hameedur  and Jalal, Ahmad },       
TITLE={Unmanned aerial vehicles for human detection and recognition using neural-network model},        
JOURNAL={Frontiers in Neurorobotics},        
VOLUME={18},
MONTH = {Dec. 04,},
YEAR={2024}, 
ISSN={1662-5218}}

@Article{mueller2024indoorhumanfollow,
AUTHOR = {Gómez, Juan and Aycard, Olivier and Baber, Junaid},
TITLE = {Efficient Detection and Tracking of Human Using 3D LiDAR Sensor},
JOURNAL = {Sensors},
VOLUME = {23},
MONTH = {May 12,},
YEAR = {2023},
NUMBER = {10},

}

@Article{gaigalas2025astarvision,
AUTHOR = {Meng, Wenlong and Zhang, Xuegang and Zhou, Lvzhuoyu and Guo, Hangyu and Hu, Xin},
TITLE = {Advances in UAV Path Planning: A Comprehensive Review of Methods, Challenges, and Future Directions},
JOURNAL = {Drones},
VOLUME = {9},
NUMBER = {5},
MONTH = {May 16,},
YEAR = {2025},

}

@Article{baidya2024landing,
AUTHOR = {Li, Jian and Liao, Changyi and Zhang, Weijian and Fu, Haitao and Fu, Shengliang},
TITLE = {UAV Path Planning Model Based on R5DOS Model Improved A-Star Algorithm},
JOURNAL = {Applied Sciences},
VOLUME = {12},
MONTH = {Nov. 08,},
YEAR = {2022},
NUMBER = {22},
}

@INPROCEEDINGS{kucukerdem2025autocontrol,
  author={Zhou, Qiang and Liu, Guangcai},
  booktitle={Proc. IEEE Int. Conf. on Unmanned Systems (ICUS)}, 
  title={UAV Path Planning Based on the Combination of A-star Algorithm and RRT-star Algorithm}, 
  year={2022},
  volume={},
  number={},
month = {Oct. 28-30,},
  pages={146--151}}

@ARTICLE{ego_planner,
  author={Zhou, Xin and Wang, Zhepei and Ye, Hongkai and Xu, Chao and Gao, Fei},
  journal={IEEE Robotics and Automation Letters}, 
  title={EGO-Planner: An ESDF-Free Gradient-Based Local Planner for Quadrotors}, 
  year={2021},
  volume={6},
  number={2},
month = {Dec. 28},
  pages={478--485}}

@INPROCEEDINGS{liang2024dtg,
  author={Liang, Jing and Payandeh, Amirreza and Song, Daeun and Xiao, Xuesu and Manocha, Dinesh},
  booktitle={Proc. IEEE Int. Conf. on Intelligent Robots and Systems (IROS)}, 
  title={DTG : Diffusion-based Trajectory Generation for Mapless Global Navigation}, 
  year={2024},
  volume={},
  number={},
month = {Oct. 14-18,},
  pages={5340--5347}}

@INPROCEEDINGS{vuckovic2024diffnav,
  author={Sridhar, Ajay and Shah, Dhruv and Glossop, Catherine and Levine, Sergey},
  booktitle={Proc. IEEE Int. Conf. Conference on Robotics and Automation (ICRA)}, 
  title={NoMaD: Goal Masked Diffusion Policies for Navigation and Exploration}, 
  year={2024},
  volume={},
  number={},
month = {May 13-17,},
  pages={63--70}}

@ARTICLE{wolf2025manipsurvey,
AUTHOR={Wolf, Rosa  and Shi, Yitian  and Liu, Sheng  and Rayyes, Rania },           
TITLE={Diffusion models for robotic manipulation: a survey},          
JOURNAL={Frontiers in Robotics and AI},          
VOLUME={12},  
MONTH = {Sept. 09,},
YEAR={2025}}

@misc{ubukata2024planningreview,
      title={Diffusion Model for Planning: A Systematic Literature Review}, 
      author={Toshihide Ubukata and Jialong Li and Kenji Tei},
      year={2024},
      note={arxiv:2408.10266}, 
}

@INPROCEEDINGS{uavvlrr2025,
  author={Yaqoot, Yasheerah and Mustafa, Muhammad Ahsan and Sautenkov, Oleg and Lykov, Artem and Serpiva, Valerii and Tsetserukou, Dzmitry},
  booktitle={IEEE Intelligent Vehicles Symposium (IV)}, 
  title={UAV-VLRR: Vision-Language Informed NMPC for Rapid Response in UAV Search and Rescue}, 
  year={2025},
  volume={},
  number={},
 month = {June 22-25},
  pages={1195-1200}}

@misc{flightgpt2025,
      title={FlightGPT: Towards Generalizable and Interpretable UAV Vision-and-Language Navigation with Vision-Language Models}, 
      author={Hengxing Cai and Jinhan Dong and Jingjun Tan and Jingcheng Deng and Sihang Li and Zhifeng Gao and Haidong Wang and Zicheng Su and Agachai Sumalee and Renxin Zhong},
      year={2025},
     note={arxiv:2505.12835}, 
}

@ARTICLE{navblip2025, 
AUTHOR={Li, Ye  and Yang, Li  and Yang, Meifang  and Yan, Fei  and Liu, Tonghua  and Guo, Chensi  and Chen, Rufeng },          
TITLE={NavBLIP: a visual-language model for enhancing unmanned aerial vehicles navigation and object detection},        
JOURNAL={Frontiers in Neurorobotics},       
VOLUME={Volume 18 - 2024},
MONTH = {Jan. 24,},
YEAR={2025},
}

@misc{openvocabvlm2025,
      title={Grounded Vision-Language Navigation for UAVs with Open-Vocabulary Goal Understanding}, 
      author={Yuhang Zhang and Haosheng Yu and Jiaping Xiao and Mir Feroskhan},
      year={2025},
     note={arxiv:2506.10756}, 
}

@Article{aienhanced2025,
AUTHOR = {Gaitan, Nicoleta Cristina and Batinas, Bianca Ioana and Ursu, Calin},
TITLE = {AI-Enhanced Rescue Drone with Multi-Modal Vision and Cognitive Agentic Architecture},
JOURNAL = {AI},
VOLUME = {6},
MONTH = {Oct. 16,},
YEAR = {2025},
NUMBER = {10},
}

@INPROCEEDINGS{Rollo_2023,
  author={Rollo, Federico and Zunino, Andrea and Raiola, Gennaro and Amadio, Fabio and Ajoudani, Arash and Tsagarakis, Nikolaos},
  booktitle={Proc. IEEE Int. Conf. on Advanced Robotics and Its Social Impacts (ARSO)}, 
  title={FollowMe: a Robust Person Following Framework Based on Visual Re-Identification and Gestures}, 
  year={2023},
  volume={},
  number={},
  month = {June 05-07,},
  pages={84--89}}

@INPROCEEDINGS{zhou2019omniscalefeaturelearningperson,
  author={Zhou, Kaiyang and Yang, Yongxin and Cavallaro, Andrea and Xiang, Tao},
  booktitle={Proc. IEEE Int. Conf. on Computer Vision (ICCV)}, 
  title={Omni-Scale Feature Learning for Person Re-Identification}, 
  year={2019},
  volume={},
  number={},
month = {Oct. 27-29,},
  pages={3701--3711}}

@misc{KennyLHW_VLNGo2Matterport_2024,
  author       = {Kenny LHW},
  title        = {VLN-Go2-Matterport Dataset},
  year         = {2024},
  howpublished = {\url{https://huggingface.co/datasets/Kennylhw/VLN-Go2-Matterport}},
  note         = {Accessed: 2025-02-08}
}

@INPROCEEDINGS{UAV_VLA,
  author={Sautenkov, Oleg and Yaqoot, Yasheerah and Lykov, Artem and Mustafa, Muhammad Ahsan and Tadevosyan, Grik and Akhmetkazy, Aibek and Cabrera, Miguel Altamirano and Martynov, Mikhail and Karaf, Sausar and Tsetserukou, Dzmitry},
  booktitle={ACM/IEEE Int. Conf. on Human-Robot Interaction (HRI)}, 
  title={UAV-VLA: Vision-Language-Action System for Large Scale Aerial Mission Generation}, 
  month = {March 4-6,},
  year={2025},
  volume={},
  number={},
  pages={1588--1592}}

@misc{swarmdiffusionen,
      title={SwarmDiffusion: End-To-End Traversability-Guided Diffusion for Embodiment-Agnostic Navigation of Heterogeneous Robots}, 
      author={Iana Zhura and Sausar Karaf and Faryal Batool and Nipun Dhananjaya Weerakkodi Mudalige and Valerii Serpiva and Ali Alridha Abdulkarim and Aleksey Fedoseev and Didar Seyidov and Hajira Amjad and Dzmitry Tsetserukou},
      year={2025},
      note={arxiv:2512.02851}, 
}

@INPROCEEDINGS{agile_pilot,
  author={Khan, Roohan Ahmed and Serpiva, Valerii and Aschalew, Demetros and Fedoseev, Aleksey and Tsetserukou, Dzmitry},
  booktitle={Proc. IEEE Int. Conf. on Unmanned Aircraft Systems (ICUAS)}, 
  title={AgilePilot: DRL-Based Drone Agent for Real-Time Motion Planning in Dynamic Environments by Leveraging Object Detection}, 
  month = {May 14-17,},
  year={2025},
  volume={},
  number={},
  pages={185--192}}

@INPROCEEDINGS{marlander,
  author={Aschu, Demetros and Peter, Robinroy and Karaf, Sausar and Fedoseev, Aleksey and Tsetserukou, Dzmitry},
  booktitle={Proc. IEEE Int. Conf. on Systems, Man, and Cybernetics (SMC)}, 
  title={MARLander: A Local Path Planning for Drone Swarms using Multiagent Deep Reinforcement Learning}, 
  month = {Oct. 6-10,},
  year={2024},
  volume={},
  number={},
  pages={2943--2948}}

@INPROCEEDINGS{DiPPeR,
  author={Liu, Jianwei and Stamatopoulou, Maria and Kanoulas, Dimitrios},
  booktitle={2024 IEEE International Conference on Robotics and Automation (ICRA)}, 
  title={DiPPeR: Diffusion-based 2D Path Planner applied on Legged Robots}, 
  month = {May 13-17,},
  year={2024},
  volume={},
  number={},
  pages={9264--9270}}
\end{document}